\documentclass[11pt,a4paper]{article}
\usepackage{authblk}
\usepackage[hyperref]{acl2017}
\usepackage{times}
\usepackage{multirow}
\usepackage{latexsym}
\usepackage{xcolor}
\usepackage{tabularx}
\usepackage[normalem]{ulem}
\usepackage{url}
\usepackage{breakurl}
\usepackage{caption}
\usepackage{pgfplots}
\usepackage{tikz}
\usetikzlibrary{patterns}
\usepackage{subfig}

\pgfplotsset{compat=1.11,
    /pgfplots/ybar legend/.style={
    /pgfplots/legend image code/.code={%
       \draw[##1,/tikz/.cd,yshift=-0.25em]
        (0cm,0cm) rectangle (3pt,0.8em);},
   },
}

\definecolor{forestgreen}{rgb}{0.13, 0.55, 0.13}
\definecolor{bblue}{HTML}{4F81BD}
\definecolor{rred}{HTML}{C0504D}
\definecolor{ggreen}{HTML}{9BBB59}
\definecolor{ppurple}{HTML}{9F4C7C}

\aclfinalcopy 

\newcommand{\secref}[1]{Section~\ref{ssec:#1}}
\newcommand{\figref}[1]{Figure~\ref{#1}}
\newcommand{\tabref}[1]{Table~\ref{#1}}
\newcommand{\isection}[2]{\section{#1}\label{ssec:#2}}
\newcommand{\isectionb}[1]{\section{#1}\label{ssec:#1}}
\newcommand{\isubsectionb}[1]{\subsection{#1}\label{ssec:#1}}

\title{The Effect of Different Writing Tasks on Linguistic Style:\\ A Case Study of the ROC Story Cloze Task}
\author[1,2]{\bf Roy Schwartz}
\author[1]{\bf Maarten Sap}
\author[1]{\bf Ioannis Konstas}
\author[1]{\\ \bf Li Zilles}
\author[1]{\bf Yejin Choi}
\author[1]{\bf Noah A.~Smith}
\affil[1]{Paul G.~Allen School of Computer Science \& Engineering,\protect\\University of Washington, Seattle, WA, USA}
\affil[2]{Allen Institute for Artificial Intelligence, Seattle, WA, USA}
\affil[  ]{\tt \{roysch,msap,ikonstas,lzilles,yejin,nasmith\}@cs.washington.edu}

\date{}

\begin{document}
\maketitle
\begin{abstract}
A writer's style depends not just on personal traits but also on her intent and mental state.
In this paper, we show how variants of the same writing task can lead to measurable differences in writing style.
We present a case study based on 
the  {\it story cloze task} \cite{Mostafazadeh:2016},
where annotators were assigned similar writing tasks with different constraints: (1) writing an entire story, (2) adding a story ending for a given story context, and (3) adding an incoherent ending to a story.
We show that a simple linear classifier informed by stylistic features is able to successfully distinguish among the three cases, without even looking at the story context.
In addition, combining our stylistic features with language model predictions reaches state of the art performance on the story cloze challenge.
Our results demonstrate that different task framings can dramatically affect the way people write.\footnote{This paper extends our LSDSem 2017 shared task submission \cite{Schwartz:2017b}.}

\end{abstract}

\section{Introduction}
Writing style is expressed through a range of linguistic elements such as words, sentence structure, and rhetorical devices.
It is influenced by personal factors such as age and gender \cite{Schler:2006}, 
by personality traits such as agreeableness and openness  \cite{Ireland:2014b},
as well as by mental states such as sentiment \cite{Davidov:2010}, sarcasm \cite{Tsur:2010}, and deception \cite{Feng:2012}.  
In this paper, we study the extent to which writing style is affected by the nature of the writing task the writer was asked to perform, since
different tasks likely engage different cognitive processes \cite{Campbell:2003,Banerjee:2014}.\footnote{For the purposes of this paper, \emph{style} is defined as content-agnostic writing characteristics, such as the number of words in a sentence.}

\begin{table}[!t]
\begin{tabularx}{\linewidth}
{l>{\setlength\hsize{0.67\hsize}}X
>{\setlength\hsize{1.33\hsize}}X}
{\bf Story Prefix} & {\bf Ending} \\ \hline
\multirow{4}{*}{\parbox{3.5cm}{\vspace{.1cm}John liked a girl at his work.	He tried to get her attention by acting silly.	She told him to grow up. John confesses he was trying to make her like him more.}} & \vspace{-0.2cm}	{\color{blue}{{\bf She feels flattered and asks John on a date.}}} 	 \\\cline{2-2}

& 	{\color{forestgreen}{\vspace{-0.3cm} \it The girl found this charming, and gave him a second chance.\vspace{-0.4cm}
}} \\ \cline{2-2}
& {\color{red}{
\vspace{-0.3cm}\uline{John was happy about being rejected.}
}} \\\hline 
\end{tabularx}
\caption{\label{ROC-example}
Examples of stories from the story cloze task. The table shows a story prefix with three contrastive endings:
The {\color{blue}{\bf original}} ending, a {\color{forestgreen}{\it coherent}} ending and a {\color{red}{\uline{incoherent}}} one.
}
\end{table}

We show that similar writing tasks with different constraints on the
author can lead to measurable differences in her writing style.
As a case study, we present experiments   based on 
the recently introduced ROC story cloze task \cite{Mostafazadeh:2016}. 
In this task, authors were asked to write five-sentence self-contained stories, henceforth {\it original} stories.
Then, 
each original story was given to a different author, 
who was shown only the first four sentences as a story context, 
and asked to write two contrasting story endings: a {\it right} (coherent) ending, and a {\it wrong} (incoherent) ending. 
Framed as a story cloze task, the goal of this dataset is to serve as a commonsense challenge for NLP and AI research. 
\tabref{ROC-example} shows an example of an {original} story, a {coherent} story, and an {incoherent} story.

While the story cloze task was originally designed to be a story understanding challenge, 
its annotation process introduced three variants of the same writing task: writing an {\it original}, {\it right}, or {\it wrong} ending to a short story.
In this paper, we show that a linear classifier informed by stylistic features can distinguish among the different endings to a large degree, even without looking at the story context (64.5--75.6\% binary classification results).

Our results allow us to make a few key observations.
First, people adopt a different writing style when asked to write coherent vs.~incoherent story endings.
Second,  people change their writing style when writing the entire story on their own compared to 
writing only the final sentence for a given story context written by someone else.

In order to further validate our method, we also directly tackle the
story cloze task. 
Adapting our classifier to the task, we obtain 72.4\% accuracy, only 2.3\% below state of the art results.
We also show that the style differences captured by our model can be combined with neural language models to make a better use of the story context. 
Our final model that combines context with stylistic features achieves
a new state of the art---75.2\%---an additional 2.8\% gain.

The contributions of our study are threefold. 
First, findings from our study can potentially shed light on 
how different kinds of cognitive load influence the style of written language. 
Second, combined with recent similar findings of \citet{Cai:2017}, our results indicate that when designing new NLP tasks, special attention needs to be paid to the instructions given to authors.
Third, we establish a new state of the art result on the commonsense story cloze challenge. 
Our code is available at \url{https://github.com/roys174/writing_style}.

\isection{Background: The Story Cloze Task}{ROC_Story}
To understand how different writing tasks affect writing style, 
we focus on the \textit{story cloze task} \cite{Mostafazadeh:2016}. 
While this task was developed to facilitate representation and learning of commonsense story understanding,
its design included a few key choices which  make it ideal for our study. 
We describe the task below.

\paragraph{ROC stories.}

The ROC story corpus consists of 49,255 five-sentence 
stories, collected on Amazon Mechanical Turk (AMT).\footnote{Recently,
  additional 53K stories were released, which results in roughly
  100K stories.}
Workers were instructed to write a coherent self-contained story, which has a clear beginning and end. 
To collect a broad spectrum of commonsense knowledge, there was no imposed subject for the stories,
which resulted in a wide range of different topics.

\paragraph{Story cloze task.}
After compiling the story corpus, the {\it story cloze task}---a task based on the corpus---was introduced.
A subset of the stories was selected, and only the first four sentences of each story were presented to AMT workers.
Workers were asked to write a pair of new story endings for each story context: one {\it right} and one {\it wrong}.
Both endings were required to complete the story using one of the characters in the story context. 
Additionally,  the endings were required to be ``realistic and sensible'' \cite{Mostafazadeh:2016} when read out of context.

The resulting stories, both {\it right} and {\it wrong}, were then individually rated for coherence and meaningfulness by additional AMT workers.
Only stories rated as simultaneously coherent with a {\it right} ending and neutral with a {\it wrong} ending were selected for the task. 
It is worth noting that workers rated the stories as a whole, not only the endings.

Based on the new stories, \citet{Mostafazadeh:2016} proposed the {\it story cloze task}. 
The task is simple:  given a pair of stories that differ only in their endings, the system decides which ending is {\it right} and which is {\it wrong}. 
The official training data contains only the original stories (without alternative endings), while development and test data consist of the revised stories with alternative endings (for a different set of original stories that are not included in the training set).
The task was suggested as an extensive evaluation framework:
as a commonsense story understanding task, 
as the shared task for the  Linking Models of Lexical, Sentential and Discourse-level Semantics workshop (LSDSem 2017, \citealp{LSDSem:2017}), and as a testbed for vector-space evaluation \cite{mostafazadeh2016story}.

Interestingly, only very recently, one year after the task was first introduced, the published benchmark on this task surpassed 60\%.
This comes in contrast to other recent similar machine reading tasks such as CNN/DailyMail \cite{hermann2015teaching}, SNLI \cite{bowman2015large}, LAMBADA \cite{Paperno:2016} and SQuAD \cite{rajpurkar2016squad}, for which results improved dramatically over similar or much shorter periods of time. This suggests that this task is challenging and that high performance is hard to achieve.

In addition, \citet{Mostafazadeh:2016} made substantial efforts to ensure the quality of this dataset. 
First, each pair of endings was written by the same author, which
ensured that style differences between authors could not be used to solve the task. 
Furthermore, Mostafazadeh et al.~implemented nine baselines for the task, using surface level features as well as narrative-informed ones, and showed that each of them reached roughly chance-level.
These results suggest that real understanding of text is required in order to solve the task.
In this paper, we show that this is not necessarily the case, by demonstrating that a simple linear classifier informed with style features reaches near state of the art results on the task---72.4\%.

\paragraph{Different writing tasks in the story cloze task.}
Several key design decisions make the task an interesting testbed for the purpose of this study.
First, the training set for the task (ROC Stories corpus) is not a
training set in the usual sense,\footnote{I.e., the training
  instances are not drawn from a population similar to the one that future
  testing instances will be drawn from.
}  as it contains only positive ({\it right}) examples, and not negative ({\it wrong}) ones. 

On top of that, the {\it original} endings, which serve as positive training examples, were generated differently from the {\it right} endings, which serve as the positive examples in the development and test sets. 
While the former are part of a single coherent story written by the same author, the latter were generated by letting an author read four sentences, 
and then asking her to generate a fifth {\it right} ending. 

Finally, although the {\it right} and {\it wrong} sentences were generated by the same author, 
the tasks for generating them were quite different: in one case, the author was asked to write a {\it right} ending, which would create a coherent five-sentence story along with the other four sentences. In the other case, the author was asked to write a {\it wrong} ending, which would result in an incoherent five-sentence story. 

\isection{Surface Analysis of the Story Cloze Task}{Surface}
We begin by computing several characteristics of the three types of endings: {\it original} endings (from the ROC story corpus training set), {\it right} endings and {\it wrong} endings (both from the story cloze task development set).
Our analysis  reveals several style differences between different groups. 
First, {\it original} endings are on average longer (11 words per
sentence) than {\it right} endings (8.75 words), which are in turn
slightly longer than {\it wrong} ones (8.47 words). 
The latter finding is consistent with previous work, which has shown that sentence length is also indicative of whether a text was deceptive \cite{qin2004exploratory,yancheva2013automatic}. 
Although writing {\it wrong} sentences is not the same as deceiving, it is not entirely surprising to observe similar trends in both tasks.

\begin{figure*}[t!]
\begin{small}
\subfloat[POS tags]{

\begin{tikzpicture}
    \begin{axis}[
        width  = 0.4*\textwidth,
        height = 4cm,
        ybar=\pgflinewidth,
        x=1.25cm,
       bar width=5pt,
        ymajorgrids = true,
        ylabel = {Frequency in Corpus (\%)},
        symbolic x coords={NN, VBD, PRP, DT, NNP},
        xtick = data,
        scaled y ticks = false,
        ymin=3,
    ]
    \addplot[style={blue,fill=blue,pattern=horizontal lines, pattern color=blue,mark=none}]
            coordinates {(NN, 15.3) (VBD, 13.4) (PRP, 10.1) (DT, 9.1) (NNP, 4.7)};

        \addplot[style={forestgreen,fill=forestgreen,pattern=north west lines, pattern color=forestgreen,mark=none}]
            coordinates {(NN, 15.7) (VBD, 15.8) (PRP, 9.7) (DT, 9.7) (NNP, 8.3)};

        \addplot[style={red,fill=red,mark=none}]
            coordinates {(NN, 16.3) (VBD, 15.1) (PRP, 7.4) (DT, 9.4) (NNP, 9.5)};
    \end{axis}
\end{tikzpicture}
\label{roc_pos_distribution}
}
\qquad\qquad\qquad
\subfloat[Words]{
\begin{tikzpicture}
    \begin{axis}[
        width  = 0.4*\textwidth,
        height = 4cm,
        ybar=\pgflinewidth,
        x=1.25cm,
       bar width=5pt,
        ymajorgrids = true,
    symbolic x coords={to, and, I, hates, !},
        xtick = data,
        scaled y ticks = false,
        ymin=0,
    ]
 
   \addplot[style={blue,fill=blue,pattern=horizontal lines, pattern color=blue,mark=none}]
            coordinates {(to, 3) (and, 2.5) (I, 1) (hates, 0) (!, 0.9)};

        \addplot[style={forestgreen,fill=forestgreen,pattern=north west lines, pattern color=forestgreen,mark=none}]
            coordinates {(to, 3.3) (and, 1.8) (I, 1.2) (hates, 0) (!, 0.2)};

        \addplot[style={red,fill=red,mark=none}]
            coordinates {(to, 3.5) (and, 1.3) (I, 1.4) (hates, 0.6) (!, 0.1)};
   
    \end{axis}
\end{tikzpicture}
\label{roc_word_distribution}
}
\end{small}
\caption{The distribution of five frequent POS tags
  (\ref{roc_pos_distribution}) and words (\ref{roc_word_distribution})
  across {\color{blue}{\emph{original}}} endings (horizontal lines) from the story cloze training set, and
  {\color{forestgreen}{\emph{right}}} (diagonal lines) and {\color{red}{\emph{wrong}}} (solid lines) endings, both from the story cloze task development set.}

\end{figure*}
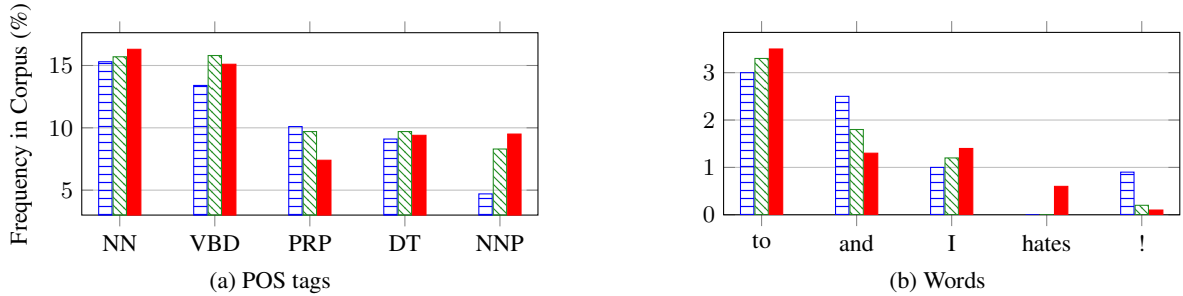

Second, \figref{roc_pos_distribution} shows the distribution of five frequent POS tags in all three groups. 
The figure shows that both {\it original} and {\it right} endings use pronouns more frequently than {\it wrong} endings.
Once again, deceptive text is also characterized by fewer pronouns compared to truthful text \cite{Newman:2003}.

Finally, \figref{roc_word_distribution} presents the distribution of five frequent words across the different groups. 
The figure shows that {\it original} endings use coordinations (``and'') more than  {\it right} endings, and substantially more than {\it wrong} ones. 
Furthermore, {\it original} and {\it right} endings seem to prefer
enthusiastic language (e.g., ``!''), while {\it wrong} endings tend to
use more negative language (``hates''), similar to deceptive text \cite{Newman:2003}.
Next we show that these style differences are not anecdotal, but can
be used to distinguish among the different types of story endings.

\isectionb{Model}

To what extent do different writing constraints lead authors to adopt different writing styles?
In order to answer this question, we first use simple methods that have been shown to be very effective for recognizing style (see \secref{Related}).
We describe our model below.

We train a logistic regression classifier to categorize an ending,
either as {\it right} vs.~{\it wrong} or as {\it original} vs.~{\it new} ({\it right}).
Each feature vector is computed using the words in one ending, without considering earlier parts of the story. 
We use the following style features.

\begin{itemize}
\item\textit{\textbf{Length}:} the number of words in the sentence.
\item\textit{\textbf{Word $n$-grams:}} we use sequences of 1--5
  words. Following \citet{Tsur:2010} and \citet{Schwartz:2013}, we distinguish between high frequency and low frequency words. 
Specifically, we replace content words (nouns, verbs, adjectives, and adverbs), which are often low frequency, with their part-of-speech tags.
\item\textit{\textbf{Character $n$-grams:}} character $n$-grams are one of the most useful features in identifying author style \cite{Stamatatos:2009}. 
We use character $4$-grams.\footnote{Experiments with $5$-grams on our development set reached similar performance.}
\end{itemize}

\isectionb{Experiments}
We design two experiments to answer our research questions. 
The first is an attempt to distinguish between {\it right} and {\it wrong} endings,
the second  between {\it original} endings and {\it new} ({\it right}) endings.
For completeness, we also run a third experiment, which compares between {\it original} and {\it wrong} endings.

\paragraph{Experiment 1: right/wrong endings.}
The goal of this experiment is to measure the extent to which  style features capture differences between the {\it right} and {\it wrong} endings.
As the story cloze task doesn't have a training corpus for the {\it
  right} and {\it wrong} endings (see \secref{ROC_Story}), we use the
development set as our training set, holding out 10\% for development
(3,366 training endings, 374 for development). 
 We keep the story cloze test set as is (3,742 endings).

It is worth noting that our classification task is slightly different from the story cloze task. 
Instead of classifying pairs of endings, one which is {\it right} and
another which is {\it wrong}, our classifier decides about each ending
individually, whether it is \emph{right} (positive instance) or
\emph{wrong} (negative instance).
By ignoring the coupling between {\it right} and {\it wrong} pairs, 
we are able to decrease the impact of author-specific style differences,
and focus on the difference between the styles accompanied with {\it right} and {\it wrong} writings.

\paragraph{Experiment 2: original/new endings.}

Here the goal is to measure whether writing the ending as part of a
story imposes different style compared to writing a {\it new} ({\it right}) ending to an existing story.
We use the endings of the ROC stories as our {\it original} examples and {\it right} endings from the story cloze task  as {\it new} examples.
As there are far more {\it original} instances than {\it new}
instances, we randomly select five  {\it original} sets, each with the same number of
instances as we have
\emph{new} instances (3,366 training endings, 374 development endings, and 3,742 test endings).
We train five classifiers, one with each of the {\it original} training sets, and report the average classification result.

\paragraph{Experiment 3: original/wrong endings.}
For completeness, we measure the extent to which our classifier can discriminate between {\it original} and {\it wrong} endings. We replicate Experiment 2, this time replacing {\it right} endings with {\it wrong} ones.

\paragraph{Experimental setup.}
In all experiments, we add a \textsc{start} symbol at the beginning
of each sentence.\footnote{99\% of all sentences end with a period
  or an exclamation mark, so we do not add a \textsc{stop} symbol.}
For computing our features, we keep $n$-gram (character or word) features that occur at least five times in the training set.
All feature values are normalized to $[0, 1]$.
For the POS features, we tag all endings with the Spacy POS tagger.\footnote{\url{http://spacy.io/}}
We use  Python's sklearn logistic regression implementation \cite{scikit-learn} with $L_2$
regularization, performing grid search on the development set to
tune a single hyperparameter---the regularization parameter.

\isubsectionb{Results}
\tabref{results} shows our results.  In all experiments,
our model achieves performance well above what would be expected under
chance (50\% by design).  Noting again that our model ignores the
story context (the preceding four sentences), our model is unable to
capture any notion of coherence.   This finding provides
strong evidence that the authors' style was affected by the writing task they
were given to perform.

\begin{table}[!t]
\begin{center}
\begin{tabular}{|l|c|} \hline
{\bf Experiment} & {\bf Accuracy} \\ \hline
{\sl right vs. wrong} & 0.645 \\ \hline
{\sl original vs. right} & 0.685 \\ \hline
{\sl original vs. wrong} & 0.756 \\ \hline
\end{tabular}
\end{center}
\caption{\label{results}Results of  experiments 1 ({\it right vs.~wrong}), 2 ({\it original vs.~right (new)}) and 3 ({\it
  original vs.~wrong (new)} endings).
In all cases, our setup implies a 50\% random baseline.}
\end{table}

\subsection{Story Cloze Task}
The results of Experiment 1 indicate that {\it right} and {\it wrong} endings are characterized by different styles.
In order to further estimate the quality of our classification results, we tackle the story cloze task using our classifier.
This classification task is more constrained than Experiment 1, as two
endings are given and the question is which is \emph{right} and which is
\emph{wrong}.
We apply the classifier from Experiment 1 as follows:
if it assigns different labels to the two given endings, we keep
them.  Otherwise, the label whose posterior probability is lower is reversed.

\tabref{cloze_results} shows our results on the story cloze test
set. Our classifier obtains 72.4\% accuracy, only 2.3\% lower than state of the art results.
Importantly, unlike previous approaches,\footnote{One exception is the  EndingsOnly system \cite{Cai:2017}, which was published in concurrence with this work, and obtains roughly the same results.} our classifier does not require the story corpus training data, and in fact doesn't even consider the first four sentences of the story in question.
These numbers further support the claim that the styles of {\it right} and {\it wrong} endings are indeed very different.

\begin{table}[!t]
\small
\begin{center}
\begin{tabular}{|l|r|} \hline
{\bf Model} & {\bf Acc.} \\ \hline
{DSSM} \cite{Mostafazadeh:2016} & 0.585 \\ 
{ukp} \cite{Bugert:2017} & 0.717\\ 
{tbmihaylov} \cite{Mihaylov:2017} & 0.724\\ 
$\dagger$EndingsOnly \cite{Cai:2017} & 0.725 \\
{cogcomp}  & 0.744 \\ 
{HIER,ENCPLOTEND,ATT} \cite{Cai:2017} & 0.747 \\\hline\hline
{RNN}		& 0.677 \\ 
$\dagger${Ours} & {0.724} \\ 
{\bf Combined (ours + RNN)} & {\bf 0.752} \\ \hline\hline
Human judgment & 1.000 \\ \hline
\end{tabular}
\end{center}
\caption{\label{cloze_results} Results on the test set of the  story cloze task.
The middle block are our results.
{\it cogcomp} results and human judgement scores are taken from \citet{Mostafazadeh:2017}.
Methods marked with ($\dagger$) do not use the story context in order to make a prediction.}
\end{table}

\paragraph{Combination with a neural language model.}
We investigate whether our model can benefit from state of the art text comprehension models, for which this task was designed. 
Specifically, we experiment with an LSTM-based \cite{hochreiter1997long} recurrent neural network language model (RNNLM; \citealp{mikolov2010recurrent}). 
Unlike the model in this paper, which only considers the story endings, this language model follows the protocol suggested by the story cloze task designers, and harnesses their ROC Stories training set, which consists of single-ending stories, 
as well as the story context for each pair of endings. 
We show that adding our features to this powerful language model
gives improvements over our classifier as well as the language
model.

We train the RNNLM using a single-layer LSTM of hidden dimension 512.
We use the ROC stories for training,\footnote{We use the extended, 100K stories corpus (see \secref{ROC_Story}).} setting aside 10\% for validation of the language model. 
We replace all words occurring less than 3 times with an 
out-of-vocabulary token, yielding a vocabulary size of  21,582.
Only during training, we apply a dropout rate of 60\% while running the LSTM over all 5 sentences of the stories. 
Using the Adam optimizer \cite{kingma2014adam} and a learning rate of
$\eta=0.001$, we train to minimize cross-entropy. 

To apply the language model to the classification problem, we select
as \emph{right} the ending with the higher value of
\begin{equation}
\frac{p_\theta(\textrm{ending} \mid
  \textrm{story})}{p_\theta(\textrm{ending})} \label{eq:ratio}
\end{equation}
The intuition is that a \emph{right} ending should be unsurprising (to
the model)
given the four preceding sentences of the story (the numerator), controlling for the
inherent surprisingness of the words in that ending (the denominator).

On its own, our neural language model performs moderately well on the story cloze test. 
Selecting endings based on $p_\theta(\textrm{ending} \mid \textrm{story})$ (i.e., the numerator of
Equation~\ref{eq:ratio}), we obtained only 55\% accuracy.   The ratio
in Equation~\ref{eq:ratio} achieves 67.7\%  (see
\tabref{cloze_results}).\footnote{Note that taking the logarithm of
  the expression in Equation~\ref{eq:ratio} gives the pointwise mutual
information between the story and the ending, under the language
model.}

We combine our linear model with the RNNLM by adding three features to
our classifier: the numerator, denominator, and ratio in
Equation~\ref{eq:ratio}, all in log space. 
We retrain our linear  model with the new feature set, and gain 2.8\%
absolute, reaching 75.2\%, a new state of the art result for the task.
These results indicate that context-ignorant style features can be used to obtain high
accuracy on the task, adding value even when context and a large
training dataset are used.

\isection{Further Analysis}{Ablation}

\subsection{Most Discriminative Feature Types}
A natural question that follows from this study is which style features are most
helpful in detecting the underlying task an author was asked
to perform. 
To answer this question, we re-ran Experiment 1 with different sub-groups of features. 
\tabref{subgroups} shows our results. Results show that  character $n$-grams are the most effective style predictors, reaching within  0.6\% of the full model, but that word $n$-grams also capture much of the signal, yielding 61.2\%, which is only 3.3\% worse than the full model. 
These findings are in line with previous work that used character $n$-grams along with other types of features to predict writing  style \cite{Schwartz:2013}.

\begin{table}[!t]
\begin{center}
\begin{tabular}{|l|c|} \hline
{\bf Feature Type} & {\bf Accuracy}\\ \hline
Word $n$-grams & 0.612 \\ \hline
Character $n$-grams & 0.639 \\ \hline
Full model & 0.645 \\ \hline

\end{tabular}
\end{center}
\caption{\label{subgroups}
Results on Experiment 1 with different subsets of features.
}
\end{table}

\subsection{Most Salient Features}
A follow-up question is which individual features contribute most to the classification process,
as these could shed light on the stylistic differences imposed by each of the writing tasks.

In order to answer this question, we consider the highest absolute
positive and negative coefficients in the logistic regression
classifier in Experiments 1 and 2, an approach widely used  as a
method of extracting the most salient features
\cite{Nguyen:2013,Burke:2013,Brooks:2013}. It is worth noting
  that its reliability is not entirely clear, since linear models like
  logistic regression can assign large coefficients to rare features \cite{Yano:2012}.
To mitigate this concern, we consider only features appearing in at least 5\% of the endings in our training set. 

\paragraph{Experiment 1.}
\tabref{exp1_features} shows the most salient features for {\it right} (coherent) and {\it wrong} (incoherent) endings in Experiment 1,
along with their corpus frequency. 
The table shows a few interesting trends. 
First, authors tend to structure their sentences differently when writing {coherent}  vs.~{incoherent} endings.
For instance, {incoherent} endings are more likely to start with a proper noun and end with a common noun, 
while coherent endings have a greater tendency to end with a past tense verb.

Second, {\it right} endings make wider use of coordination structures, as well as adjectives.
The latter might indicate that writing coherent stories inspires the authors to write more descriptive text compared to incoherent ones, 
as is the case in truthful vs.~deceptive text \cite{ott2011finding}.
Finally, we notice a few syntactic differences: {\it right} endings more often use infinitive verb structure, while {\it wrong} endings prefer gerunds (VBG).

\begin{table*}
\centering
\subfloat[Experiment 1]{
\begin{tabular}{|c|c|c||c|c|c|} \hline
\textit{\textbf{Right}} & Weight & Freq. &   \textit{\textbf{Wrong}}& Weight & Freq.  \\ \hline
`ed .' & 0.17 & {\color{white}{0}}6.5\%&  \textsc{start} NNP & 0.21& 54.8\%  \\ \hline
`and ' &  0.15 & 13.6\%  & NN . & 0.17 & 47.5\%  \\ \hline
JJ & 0.14 & 45.8\%  & NN NN . & 0.15 &  {\color{white}{0}}5.1\%  \\ \hline
to VB & 0.13 & 20.1\%& VBG & 0.11 & 10.1\%\\ \hline
`d th' & 0.12 & 10.9\%& \textsc{start} NNP VBD & 0.11 & 41.9\%\\ \hline
\end{tabular}
\label{exp1_features}
}
\quad
\subfloat[Experiment 2]{
\begin{tabular}{|c|c|c||c|c|c|} \hline
\textit{\textbf{Right}} & Weight & Freq. &   \textit{\textbf{Wrong}}& Weight & Freq.  \\ \hline
{\it length} & 0.81&  {\color{white}{.}}100.0\% & `.' & 0.74&93.0\%\\ \hline
`!' & 0.46&{\color{white}{00}}6.1\%& \textsc{start} NNP  & 0.40&39.2\% \\ \hline
NN & 0.35&{\color{white}{0}}78.9\% & \textsc{start} NNP VBD &0.23&29.0\% \\ \hline
RB & 0.34&{\color{white}{0}}44.7\%& NN . & 0.20&42.3\% \\ \hline
`,' & 0.32&{\color{white}{0}}12.7\%&  the NN . & 0.20&10.6\% \\ \hline

\end{tabular}
\label{exp2_features}
}
\caption{
The top 5 most heavily weighted features for predicting {\it right} vs.~{\it wrong} endings (\ref{exp1_features}) and  {\it original} vs.~{\it new} ({\it right}) endings (\ref{exp2_features}). 
{\it length} is the sentence length feature (see \secref{Model}).
}
\end{table*}

\paragraph{Experiment 2.}
\tabref{exp2_features} shows the same analysis for Experiment 2.
As noted in \secref{ROC_Story}, {\it original} endings tend to be much longer, which is indeed the most salient feature for them.
An interesting observation is that exclamation marks are a strong
indication for an  {\it original} ending. 
This suggests that authors are more likely to show or evoke enthusiasm when writing their own text compared to ending an existing text.

Finally, when comparing the two groups of salient features from both experiments, we find an interesting trend.
Several features, such as ``\textsc{start} NNP'' and  ``NN .'', which indicate {\it wrong} sentences in Experiment 1, are used to predict {\it new} (i.e., {\it right}) endings in Experiment 2. 
This indicates that, for instance, incoherent endings have a stronger tendency to begin with a proper noun compared to coherent endings, 
which in turn are more likely to do so than original endings. 
This partially explains why distinguishing between {\it original} and {\it wrong} endings is an easier task compared to the other pairs (\secref{Results}).

\isectionb{Discussion}

\paragraph{The effect of writing tasks on mental states.}
In this paper we have shown that different writing tasks affect a writer's writing style in easily detected ways.
Our results indicate that when authors are asked to write the last
sentence of a five-sentence story, they will use different style to
write a {\it right} ending compared to a {\it wrong} ending. We have
also shown that writing the ending as part of one's own five-sentence story is very different than reading four sentences and then writing the fifth.
Our findings hint that the nature of the writing task imposes a
different mental state on the author, which is expressed in ways that can be observed using extremely simple automatic tools. 

Previous work has shown that a writing task can affect mental state.
For instance, writing deceptive text leads to a significant cognitive
burden accompanied by a writing style that is different from truthful
text \cite{Newman:2003,Banerjee:2014}.
Writing tasks can even have a long-term effect,
as writing emotional texts was observed to benefit both physical and mental health \cite{Lepore:2002,Frattaroli:2006}. 
\citet{Campbell:2003} also showed that the health benefits of writing emotional text are accompanied by changes in writing style, mostly in the use of pronouns.

Another line of work has shown that writing style is affected by mental state.
First, an author's personality traits (e.g., depression, neuroticism, narcissism) affect her writing style \cite{schwartz2013personality,Ireland:2014b}.
Second, temporary changes, such as a romantic relationship \cite{Ireland:2011,Bowen:2016}, work collaboration \cite{Tausczik:2009,Gonzales:2009}, or negotiation \cite{Ireland:2014} may also affect writing style.
Finally, writing style can also change from one sentence to another, for instance between positive and negative text \cite{Davidov:2010} or when writing sarcastic text \cite{Tsur:2010}. 

This large body of work indicates a tight connection between writing
tasks, mental states, and variation in writing style.
This connection hints that the link discovered in this paper, between
different writing tasks and resulting variation in writing style,
involves differences in mental state. 
Additional investigation is required in order to further validate this hypothesis.

\paragraph{Design of NLP tasks.}
Our study also provides important insights for the future design of NLP tasks. 
The story cloze task was very carefully designed. Many factors, such
as topic diversity and temporal and causal relation diversity,  were controlled for \cite{Mostafazadeh:2016}. 
The authors also made sure each pair of endings was written by the
same author, partly in order to avoid author-specific style effects.
 Nonetheless, despite these efforts, several significant style
 differences can be found between the story cloze training and test set, as well as between the positive and negative labels. 
 
Our findings suggest that careful attention must be paid to instructions given to authors, especially in unnatural tasks such as writing a {\it wrong} ending. 
The COPA dataset \cite{Roemmele:2011}, which was also designed to test commonsense knowledge, explicitly addressed potential style differences in their instructions. In this task,  systems are presented with premises like {\it I put my plate in the sink}, and then decide between two alternatives, e.g.: (a) {\it I finished eating.} and (b) {\it I skipped dinner.}
Importantly, when writing the alternatives,  annotators were asked to be as brief as possible and avoid proper names, as well as  slang. 

Applying our story cloze classifier to this dataset yields 53.2\%
classification accuracy---close to a random baseline. 
While this could be partially explained by the smaller data size of the COPA dataset (1,000 examples compared to 3,742 in the story cloze task), this indicates that simple instructions may help alleviate the effects of writing style found in this paper.
Another way to avoid such effects is to have people rate naturally occurring sentences  by parameters
such as coherence (or, conversely, the level of surprise),
rather than asking them to generate new text.

\isection{Related Work}{Related}

\paragraph{Writing style.}
Writing style has been an active topic of research for decades. 
The models used to characterize style are often linear classifiers with style features such as character and word $n$-grams \cite{Stamatatos:2009,Koppel:2009}.
Previous work has shown that different authors can be grouped by their
writing style, according to factors such as age
\cite{Pennebaker:2003,Argamon:2003,Schler:2006,Rosenthal:2011,nguyen:2011:latech},
gender \cite{Argamon:2003,Schler:2006,bamman2014gender}, and native language
\cite{Koppel:2005,Tsur:2007,Bergsma:2012}.
At the extreme case, each individual author adopts a unique writing
style \cite{mosteller1963inference,pennebaker1999linguistic,Schwartz:2013}. 

The line of work that most resembles our work is the detection of deceptive text. 
Several researchers have used stylometric features to  predict deception 
\cite{Newman:2003,hancock2007lying,ott2011finding,Feng:2012}.
Some works even showed that gender affects a person's  writing style when lying \cite{Perez:2014b,Perez:2014a}.
In this work, we have shown that an even more subtle writing task---writing {coherent} and {incoherent} story endings---imposes different styles on the author.

\paragraph{Machine reading.}
The  story cloze task, which is the focus of this paper, is part of a wide set of machine reading/comprehension challenges published in the last few years.
These include datasets like bAbI \cite{Weston:2015}, SNLI \cite{bowman2015large}, CNN/DailyMail \cite{hermann2015teaching}, LAMBADA \cite{Paperno:2016} and SQuAD \cite{rajpurkar2016squad}. 
While these works have presented resources for researchers, 
it is often the case that these datasets suffer from methodological
problems caused by applying noisy automatic tools to generate them
\cite{Chen:2016}.\footnote{Similar problems have been shown in visual
  question answering datasets, where simple models that rely mostly on
  the question text perform competitively with state of the art models
  by exploiting language biases
  \cite{Zhou:2015,Jabri:2016}.}
In this paper, we have pointed to another methodological challenge in
designing machine reading tasks:  different writing tasks
used to generated the data affect writing style, confounding classification problems.

\isectionb{Conclusion}

Different writing tasks assigned to an author result in different
writing styles for that author.
We experimented with the story cloze task, which introduces two interesting comparison points: 
 the difference between writing a story on one's own and continuing someone else's story,
 and the difference between writing a coherent and an incoherent story ending.
In both cases, a simple linear model reveals measurable differences in writing styles, 
which in turn allows our final 
model to achieve state of the art results on the story cloze task.

The findings presented in this paper have  cognitive implications, as
they motivate further research on the effects that a writing
prompt has on an author's mental state, and also her concrete response.
They also provide valuable lessons for designing new NLP datasets.

\section{Acknowledgments}
The authors thank Chenhao Tan, Luke Zettlemoyer, Rik Koncel-Kedziorski, Rowan Zellers, Yangfeng Ji and several anonymous reviewers for helpful feedback.
This research was supported in part by Darpa CwC program through ARO (W911NF-15-1-0543), Samsung GRO, NSF IIS-1524371, and gifts from Google and Facebook.

\bibliography{acl2017}
\bibliographystyle{acl_natbib}

\end{document}